\documentclass[conference]{IEEEtran}

\usepackage{cite}

\ifCLASSINFOpdf
  \usepackage[pdftex]{graphicx}
  \graphicspath{{figures/}}
\else
  \usepackage[dvips]{graphicx}
  \graphicspath{{figures/}}
\fi

\usepackage[cmex10]{amsmath}
\usepackage{amssymb}

\usepackage[caption=false,font=footnotesize]{subfig}
\usepackage{fixltx2e}

\usepackage{balance}

\begin{document}

\title{Context Discovery for Model Learning\\in Partially Observable Environments}

\author{\IEEEauthorblockN{Nikolas J. Hemion}
\IEEEauthorblockA{SoftBank Robotics Europe (Aldebaran)\\
43 rue du Colonel Pierre Avia, Paris, France\\
Email: nhemion@aldebaran.com}
}

\maketitle

\begin{abstract}
The ability to learn a model is essential for the success of autonomous agents. Unfortunately, learning a model is difficult in partially observable environments, where latent environmental factors influence what the agent observes. In the absence of a supervisory training signal, autonomous agents therefore require a mechanism to autonomously discover these environmental factors, or \emph{sensorimotor contexts}.

This paper presents a method to discover sensorimotor contexts in partially observable environments, by constructing a hierarchical transition model. The method is evaluated in a simulation experiment, in which a robot learns that different rooms are characterized by different objects that are found in them.
\end{abstract}

\IEEEpeerreviewmaketitle

\section{Introduction}

This paper presents parts of our ongoing research to develop a method to let
robots autonomously learn a model of their sensorimotor interaction with the
world. Narrowly defined, model learning can be understood as providing an agent
with the means to internally simulate the world, for example to evaluate the
outcome of a potential action prior to its execution. This type of model
learning is studied in the reinforcement learning literature, where models are
typically used to reduce the time that is needed to optimize a policy
\cite{hester2012learning}. Most approaches assume that a suitable state
representation of the world is a priori known to the agent, and focus on the
problem of letting it estimate state transition probabilities and a reward
function through exploration. If however a representation is not provided in
advance, this narrow definition has to be extended: when learning a model, an
agent not only has to learn the effects of its own actions, it also has to come
up with a compact way to represent the state of its environment. Large parts of
this representation learning certainly constitute the unsupervised extraction
of feature representations from observed data, to reduce the dimensionality of
the input and to increase the robustness of the system
\cite{bengio2013representation}. But beyond unsupervised feature learning,
autonomous agents have to deal with the problem of ambiguity in their
observations and unobservable environmental influences on their actions.

This work takes inspiration from theories of perception and cognition in the
cognitive science literature \cite{friston2010free,o2001sensorimotor}, to study
the problem a naive agent is facing when trying to learn a model while having
to deal with ambiguity and latent environmental influences on the outcome of
its own actions. The focus here lies on developing a method that allows an
agent to construct a hierarchical model of its environment, as a way to
discover \emph{sensorimotor contexts}: situations in which the outcomes of the
agent's own actions are predictable. Once the agent is able to recognize such
contexts, it can learn to predict the effect of its own actions for each of the
contexts (the demonstration of which however lies out of the scope of this
paper).




\section{Discovering sensorimotor contexts to minimize prediction error}

Central to our approach, and as argued in more detail elsewhere
\cite{laflaquiere2015developmental}, is the idea that an agent should try to
discover predictable patterns of interaction in the flux of sensorimotor
observations. In the following, the argument will be briefly summarized.

In line with so-called ``predictive processing'' theories of cognition, which
have increasingly received attention in the recent cognitive science literature
\cite{friston2010free,seth2014predictive,hohwy2013predictive}, we argue that
autonomous agents need to learn to predict immediate and future sensorimotor
observations, to be able to react adaptively. A central factor driving the
learning and exploration of an agent should therefore be the goal to minimize
its error of prediction. To formalize this, consider an agent interacting with
its environment. We can describe the stream of its sensorimotor observations as
the transition probability
\begin{equation}
    \Pr( \mathbf{x}' ~ | ~ \mathbf{x}, \pi, \mathbf{e} ),
    \label{eq:state_transition_probability}
\end{equation}
expressing the probability for the agent to observe $\mathbf{x}'$ after having
observed $\mathbf{x}$, while executing the control policy $\pi$. Additionally,
we let the probability distribution depend on a latent variable $\mathbf{e}$,
which represents the current ``agent-environment configuration'': it summarizes
all external factors that influence the outcome of the agent's actions. For
example, imagine a robot with a control policy to grasp a bottle: executing
this policy will obviously only have a chance of success if there is a bottle
in reach of the robot in the current situation. As another example, imagine two
identical robots, standing in a corridor in front of two identical looking
doors, one leading to a kitchen, the other leading to a dining room. The
sequence of observations that the two robots would make when opening and
passing through the respective doors would of course be very different (one
would probably see a fridge, while the other would probably see a dining
table). This environmental influence, or sensorimotor context, is summarized in an abstract
way by the variable $\mathbf{e}$.

In practice, it is usually the designer of the robot who ensures that the robot
only executes its actions in suitable contexts. Either this is the case because
the robot never leaves a well-defined environment (such as a laboratory), or
the robot is manually provided with a detection mechanism (for example, a
``bottle detector'', in the case of the first example above). In both cases,
the model learning is implicitly turned into a supervised learning problem, as
the human expert determines a suitable context $\mathbf{e}$ for a task at hand
and designs the robot and/or its environment in such a way as to ensure that
the context remains suitable throughout every task execution.  However, when
the aim is to build truly autonomous robots, this becomes unfeasible. Instead,
we require a procedure to let robots learn in an unsupervised manner (or only
depend on supervision through natural interaction, to the extent that it is
also received by infants from their caregivers).

Thus, the agent cannot estimate the above transition probability when it does
not have access to a supervisory signal. What the agent does observe as it
tries to execute a policy $\pi$ in a number of unobservable sensorimotor contexts
$\mathbf{e}_i, i \in \{ 1, 2, \dots, N \}$ is a marginalized version of the
probability distribution
\begin{equation}
  \widehat{\Pr}( \mathbf{x}' ~ | ~ \mathbf{x}, \pi ) = \sum_{i=1}^N \Pr( \mathbf{x}' ~ | ~ \mathbf{x}, \pi, \mathbf{e}_i ),
\label{eq:marginalized_distribution}
\end{equation}
which summarizes the probability of observing $\mathbf{x}'$ after $\mathbf{x}$
while executing the policy $\pi$ across sensorimotor contexts.

Naturally, the entropy of this marginalized distribution increases with the
number of sensorimotor contexts which influence the outcome of the agent's
executed actions. This in turn means that the agent's prediction error rate
would increase, were it to make predictions of future observations solely based
on an estimate of this distribution. By implication, this means that the agent
can minimize its prediction error by demarginalizing the observed transition
probability distribution. The entropy would be minimal were the agent to
achieve a distribution that corresponded to the actual values of the latent
variable $\mathbf{e}$: as this variable represents all external factors that
influence the outcome of the agent's actions, any remaining stochasticity would
be intrinsic to the agent's own policy.

One might be inclined to think of the variable $\mathbf{e}$ as the set of all
possible states of the entire universe. But this would of course be entirely
misleading: instead, it is helpful to conceive of it as the most compact way to
encode all qualitatively different situations the agent can face, given its
sensorimotor apparatus. This is related to the concept of ``sensorimotor
contingency'' from the theory of perception proposed by O'Regan and No{\"e}
\cite{o2001sensorimotor}.

Assuming that an agent has successfully demarginalized the transition
probability distribution by constructing an internal model, it can estimate the
current state $\mathbf{e}$ by tracking the likelihood of individual hypotheses
over time: making predictions about future observations when assuming certain
states, and comparing the predictions with actual observations, to update
likelihoods correspondingly. Importantly, this implies that the distinguishing
property on which the agent relies to decide whether two observations belong to
the same sensorimotor context or not is \emph{temporal adjacency}: given a
high likelihood of a certain context $\mathbf{e}$, the agent assumes that the
next observation will also correspond to the same context. Furthermore, the
agent can \emph{test} whether two observations are part of the same
sensorimotor context by trying to produce the one observation after having
seen the other, by means of its own actions. If the agent can reliably
transition between two observations whenever it believes that a certain
sensorimotor context has a high likelihood, the agent can safely assume that
both observations are part of the same sensorimotor context and can update its model
accordingly. On the contrary, if the agent cannot produce some observation, it
can be inferred that this observation does not belong to the current
sensorimotor context.

To summarize, we can say that observations that belong to the same
sensorimotor context share the property that the agent can transition between
them using its own actions, or in other words, they are ``reachable'' from one
another for the agent. In contrast, observations that are not reachable in this
sense do not belong to the same sensorimotor context. The task to discover
sensorimotor contexts within the flux of sensorimotor observations can thus be
reformulated as one of finding sets of observations, for which the agent
observes a high probability of transitioning \emph{within} the set, but
observing transitions \emph{out of} the set only with a low probability.

\subsection{Mathematical Formalization}

We can formalize this by calling $T_{\pi}$ the transition probability matrix
corresponding to the probability distribution defined in
Eq.~\ref{eq:marginalized_distribution}, which when interpreted as an adjacency
matrix defines a graph of observations, with edges representing the probability
of transitioning between observations when the agent is executing the policy
$\pi$. The task of finding sensorimotor contexts is thus related to the problem
of finding clusters of densely connected nodes in this graph. This can be
approximately achieved through spectral clustering \cite{von2007tutorial},
using the method suggested by Ng and colleagues \cite{ng2002spectral}, in the
following way.

We first solve the eigenvalue problem for the transition probability matrix
$T_{\pi}$ to find the $k$ largest eigenvalues $\lambda_1, \lambda_2, \dots,
\lambda_k$ and associated eigenvectors $u_1, u_2, \dots, u_k$, and form the
matrix
\begin{equation}
U = [u_1~u_2~\dots~u_k] \in \mathbb{R}^{m \times k}
\label{eq:eigenvectormatrix}
\end{equation}
where $m$, is the number of discrete sensorimotor states that the agent can
observe. The matrix $U$ is then normalized such that each row has unit length,
resulting in the matrix $V$ with entries
\begin{equation}
V_{i, j} = \frac{U_{i, j}}{\sqrt{ \sum_l U_{i, l}^2 }}.
\label{eq:normalizedeigenvectormatrix}
\end{equation}

Since $V$ is computed from the observation transition probability matrix
$T_{\pi}$, each row still corresponds to one observation $\mathbf{x}$. By
treating each row in $V$ as a vector in $\mathbb{R}^{k}$, we can thus map the
observations into a $k$-dimensional ``spectral space''. As we will see below,
this mapping has the property that two observations sharing strong transition
probabilities tend to lie close together, whereas observations with low
transition probabilities between them lie further apart. This property can be
exploited to discover sets of observations that tend to co-occur by clustering
the points in this spectral space, for example using K-means. 


The idea to use graph clustering methods to partition a state graph has already
been suggested in the reinforcement learning literature
\cite{csimcsek2005identifying, mannor2004dynamic}, but with a different
motivation: the aim of these works is to discover ``subgoals'', to speed up
learning convergence in reinforcement learning (see also
\cite{barto2003recent}). The idea is that ``bottlenecks'' (such as doorways in
a navigation task) are important subgoals when discovering a policy, and they
can be characterized as state transitions with low probability between two
clusters of densely connected states. In contrast, here we are interested in
discovering densely connected clusters, with the aim to demarginalize the
transition probability and to discover sensorimotor contexts.

\subsection{Example: Four room world, fully observable case}

\begin{figure}
\includegraphics[width=\linewidth]{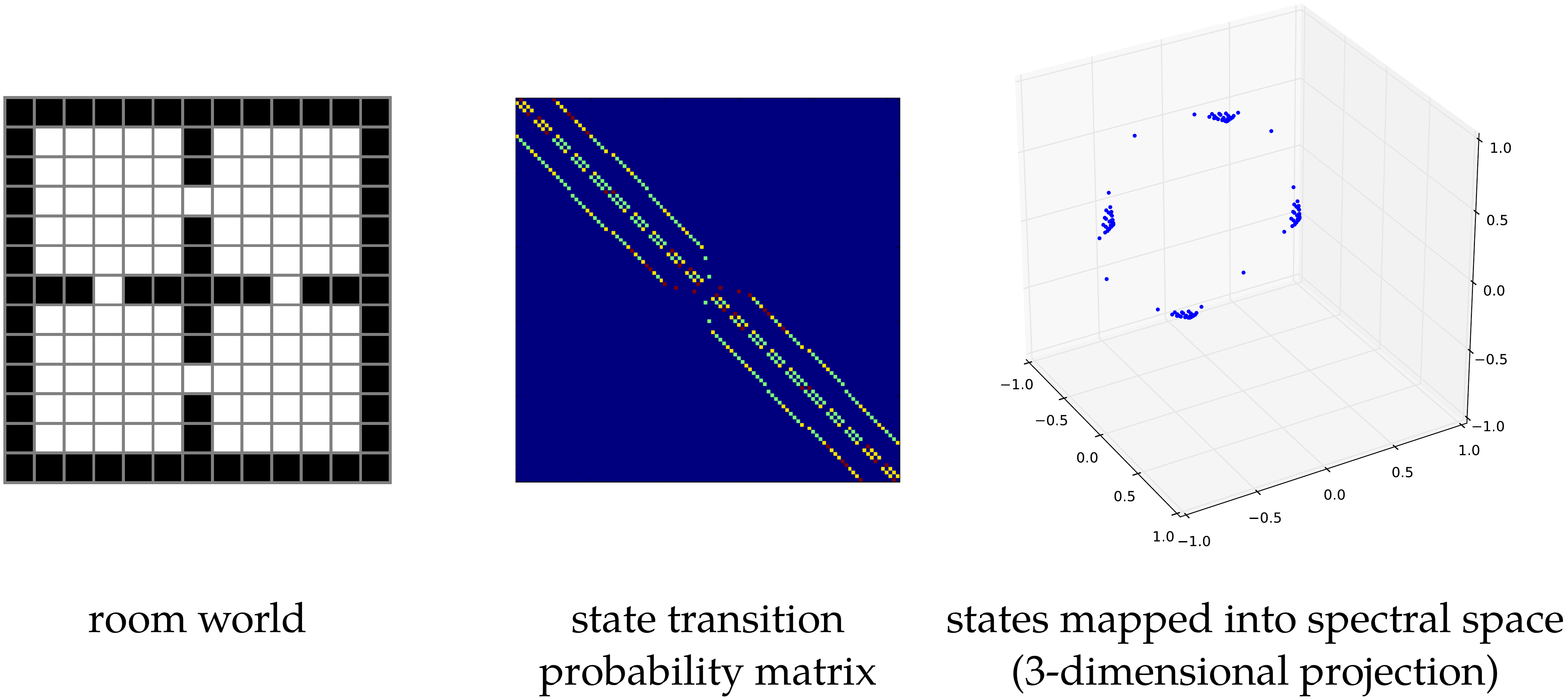}
\caption{The result of mapping the set of \emph{states} from the room
environment shown on the left into a 4-dimensional spectral space (projected
into three dimensions for visualization), based on the observed state
transition probabilities (as shown by their transition probability matrix). See
text for more details.}
\label{fig:example-1}
\end{figure}

To exemplify the mapping of observations into the spectral space, consider the
``room world'' depicted in Figure~\ref{fig:example-1}, in which an agent is
placed that is able to move up, down, left, and right. In this example, we
consider the fully observable case, meaning that the agent makes a unique
observation in each position, allowing the agent to unambiguously recognize
whenever it reaches the same position. Environments similar to this one
are often studied in the reinforcement learning literature.

As the agent explores its environment by executing an exploratory policy $\pi$
(for example a random walk), it can record the transition probabilities between
observations in a transition probability matrix $T_{\pi}$ to estimate the
marginal probability distribution in
Equation~\ref{eq:marginalized_distribution}. The resulting transition
probability matrix for the policy in which each of the agent's four actions
have equal probability to be selected is shown in Figure~\ref{fig:example-1}.

If we map the 104 unique observations (corresponding to the 104 unique
positions that the agent can be in) into a 4-dimensional spectral space using
the method described above, we obtain a distribution of observations as shown
in the figure (projected into three dimensions for visualization). We clearly
see four dense clusters, each composed of all the observations belonging to one
room. In between the clusters lie the four ``doorway'' states, through which the
agent passes when moving from one room into another.

\subsection{Example: Four room world, partially observable case}

\begin{figure}
\includegraphics[width=\linewidth]{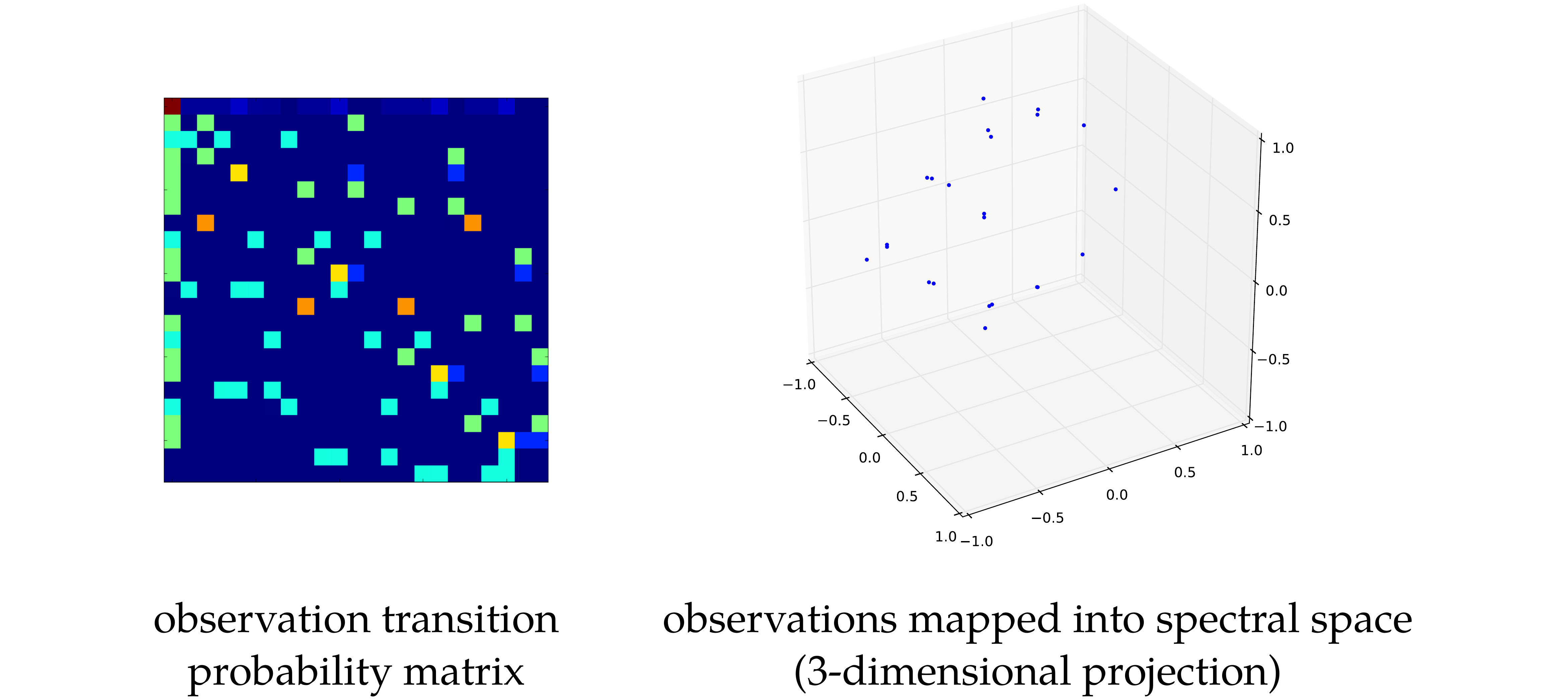}
\caption{The result of mapping the set of \emph{observations} from the room
environment shown in Figure~\ref{fig:example-1} into a 4-dimensional spectral
space (projected into three dimensions for visualization), based on the
observed observation transition probabilities (as shown by their transition
probability matrix). See text for more details.}
\label{fig:example-2}
\end{figure}

Now consider the same case in which the agent is not able to directly observe
its position in the world. Instead, we equip it with a sensor to see only the
cells surrounding its current position. This agent will not be able to
distinguish between any positions in which it makes the same observation, for
example the upper left corners of all four rooms seem identical to it. However,
to be able to predict the outcome of its own actions, it needs to know in which
room it is located: moving down from the upper left corner of the bottom right
room results in seeing the room's left door, whereas moving down from the upper
left corner of the upper left room results in seeing a wall.

Observations are ambiguous in this partially observable case. The agent's
actions can lead to different transitions between observations, depending on
the current sensorimotor context (here: which room the agent currently is in).
This ambiguity also manifests itself in the mapping into the spectral space of
the observation transition probability matrix: while some of the observations
lie closer together than others (because for example corners tend to transition
to walls, but never to other corners), the fact that the world consists of four
separate rooms is not recognizable based on the distribution in this space (in
contrast to the fully observable case).

\subsection{Dealing with partial observability by means of hierarchy}

To overcome the difficulty of partial observability, we make use of the
following intuition: while a single observation can be ambiguous, there will
always be a \emph{sequence} of observations that will unambiguously identify
any event, given the sequence is sufficiently long. For example, a single frame
from a video of a ball rolling along a plane would not allow a viewer to
determine in which direction the ball is rolling. Two consecutive frames from
the same video however are sufficient to determine the direction of the ball's
movement.

\begin{figure}
\includegraphics[width=\linewidth]{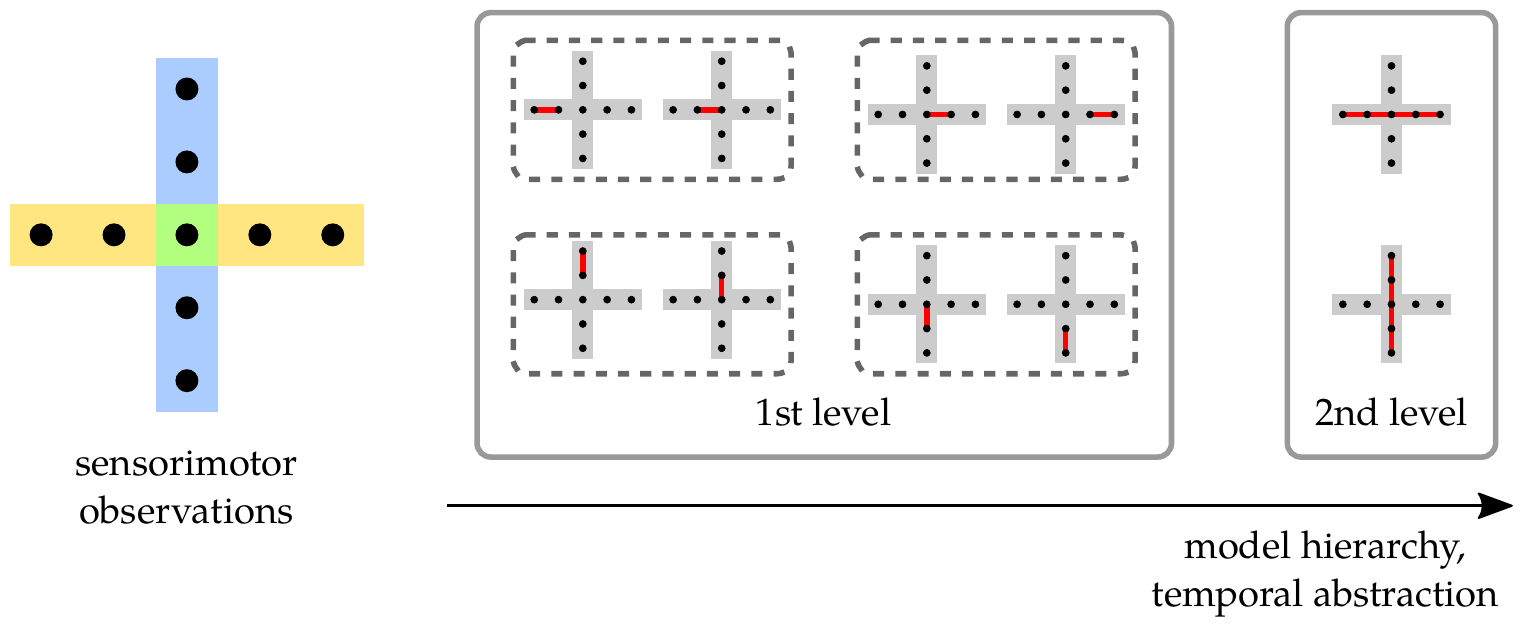}
\caption{How a model composed of a hierarchy of observation transitions extends
the temporal horizon across levels and provides an agent with a form of memory (see text for details).}
\label{fig:transition_hierarchy}
\end{figure}

What sequence length is necessary to disambiguate observations differs of
course from situation to situation -- it is not possible to fix it a priori.
Instead, the agent should be able to flexibly extend the sequence length if
necessary (i.e. when it is facing ambiguous observations). We can achieve such
a behavior by constructing a model which consists of a hierarchy of
transitions: in such a model, each successive level increases the temporal
horizon of representation. New representational levels can be constructed by
first grouping together ``similar'' transitions on the respective lower level
(ones that frequently co-occur, determined using the spectral method described
above), and subsequently determining which transitions between these
\emph{groups} of transitions are observed.

To make this idea more clear, consider the example shown in
Figure~\ref{fig:transition_hierarchy}: On the left of the figure, an abstract
example of two sensorimotor contexts is depicted, one shown in yellow, the
other shown in blue. Each context consists of five observations (shown as black
dots), most of which unambiguously belong to only one of the two contexts. Only
the observation in the center, at the ``intersection'' of the two contexts
(shown in green), can belong to either the one or the other context. Whenever
the agent makes this observation, it thus cannot determine based on the
observation alone what the current context is. In this example, a single
transition is sufficient to disambiguate contexts -- if the agent transitions
from the central observation ``vertically'', it currently is in the blue
context; otherwise, if it transitions ``horizontally'', it is in the yellow
context. The first level of the model (shown on the right of the figure)
represents transitions between observations. The agent can further create a
second layer to discover the actual latent structure of the environment, by
first grouping together ``similar'' transitions (ones that frequently co-occur;
shown as dashed boxes), and subsequently determining which transitions between
these \emph{groups} of transitions it can observe. The second level thus
represents higher-order transitions, each one corresponding to four simple
transitions.

Indeed, as illustrated by this example, the construction of such a hierarchical
model can be used to demarginalize the transition probability distribution: by
constructing an internal hierarchical model, the agent can estimate the latent
structure of the environment and obtain an internal representation of
sensorimotor contexts $\mathbf{e}$. In the next section, this will be
demonstrated by means of a simulation experiment.

\section{Simulation experiment}
\label{sec:simulation_experiment}

\begin{figure}
\includegraphics[width=\linewidth]{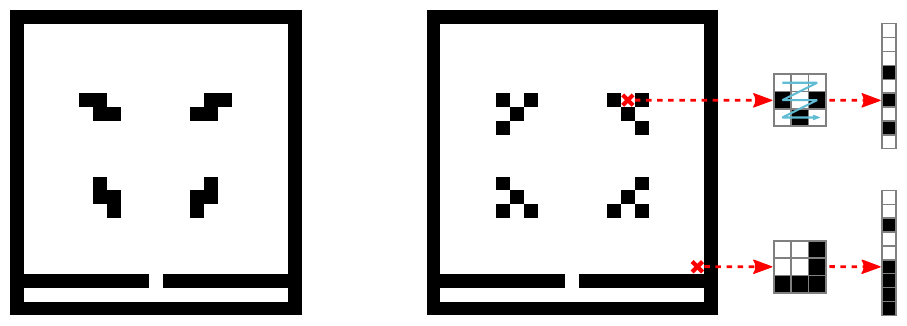}
\caption{Examples for the two types of rooms that were generated in the simulation experiment: an \emph{s}-room on the left, and a \emph{y}-room on the right. The room types differ in that they each have a unique set of ``objects'' placed inside them. A simulated robot navigates the room environment my moving up, down, to the left, or to the right. It observes only the fields immediately neighboring its current position, as a binary vector with nine entries, as shown on the right.}
\label{fig:simulation_observation}
\end{figure}

\subsection{The simulated world}

The simulation experiment should emulate the experience of a robot that
explores different rooms, without having any prior knowledge whatsoever about
the nature of rooms (neither that there exist different kinds of rooms, nor
that rooms consist of walls, that there are objects inside rooms, etc.). We
define two room types: \emph{y}-rooms and \emph{s}-rooms, each characterized by
having a unique set of ``objects'' that are placed inside them. A simulated
robot is placed in the room environment, and can navigate along a discrete grid
of positions, with the four primitive actions of moving up, down, to the left,
and to the right. At each position, the robot observes its environment in the
form of a binary occupancy vector of length 9, measuring only the grid
positions that are immediately surrounding its current position.
Figure~\ref{fig:simulation_observation} shows a visualization of the
simulation.

The robot explores its environment using a simple policy: at each time step, it
selects any of its \emph{valid} actions (ones leading to an unoccupied grid
position) with equal probability. To let the robot experience different
instances of each room, a randomization mechanism is implemented in the
simulation, in the following way. Each room has a door in its south wall,
leading to a corridor.  Whenever the robot leaves a room through this door, a
new room is randomly generated, either an \emph{s}-room or a \emph{y}-room with
equal probability. Rooms have a size of $50\times50$ tiles including walls, and
20 objects from the room's set are randomly placed in the room, ensuring that
they are not overlapping or directly adjacent to one another or to a wall.

\subsection{Simulation results}

\begin{figure*}
\includegraphics[width=\linewidth]{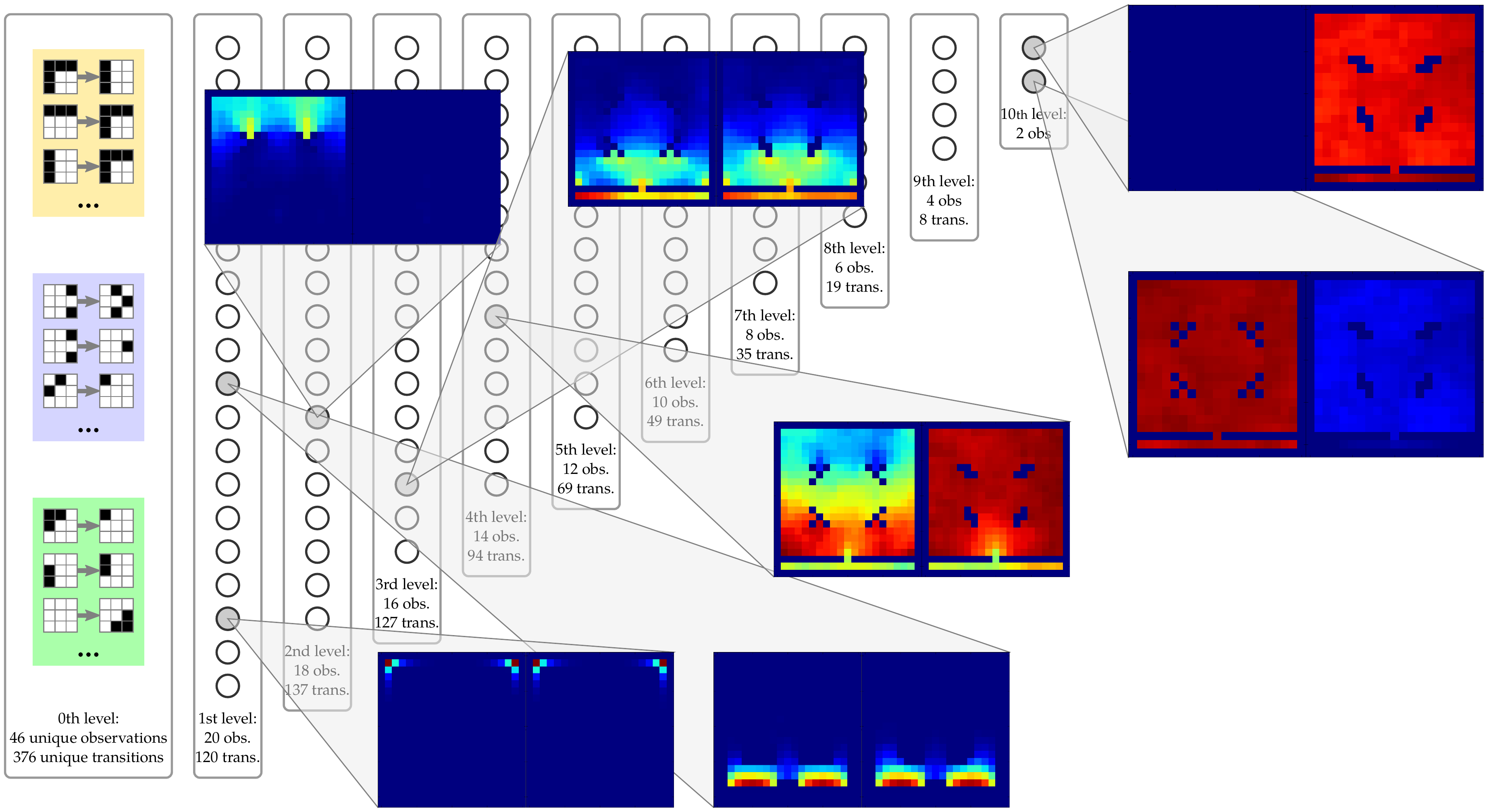}
\caption{Summary of the results of the simulation experiment. On the left (0th
level), simple transitions between observations are shown. Some of these
transitions are between room features (marked in yellow, for example moving
from a corner to a wall). Others are specific to one of the objects
(\emph{y}-type objects marked in blue, \emph{s}-type objects marked in green).
The result of applying the method for constructing a hierarchical transition
model is shown by visualizing for some nodes, where in the test rooms the agent
observes the encoded transitions. The higher the hierarchical level, the larger
the temporal abstraction of the encoded transitions is. Intermediary nodes
encode experiences such as ``entering a room from the corridor'', while the two
nodes on the highest hierarchical level encode the experiences ``being in a
\emph{y}-type room'' and ``being in an \emph{s}-type room'', respectively.}
\label{fig:Ys_results}
\end{figure*}

The robot explores its environment for a total of 10,000,000 time steps. In its
environment, the robot encounters 46 unique observations, some of which
correspond to room features (walls and corners), while others correspond to
object parts.

After the robot has completed its exploration, a hierarchical transition model
is constructed using the learning method described above, in the following way
(see also Figure~\ref{fig:Ys_results}). First, each unique transition between
observations (376 in total) is represented on the ``0th level'', and a
$376\times376$ transition probability matrix is constructed based on the
observation sequence. The mapping into its 3-dimensional spectral space is
computed, and spectral clustering in this space is performed using
agglomerative clustering, to find 20 clusters of transitions. Each of the
clusters is then treated as a ``meta-observation'' for the construction of the
next hierarchical level.

This procedure (finding all unique transitions between observations on the
respective lower level, constructing a transition probability matrix,
performing spectral clustering) is then repeated to form a total of 10
hierarchical levels, where the number of clusters is decreased by 2 on each
consecutive level, giving 18, 16, \dots, 4, 2 clusters, respectively.


To visualize the outcome of the training, two ``test rooms'' (one for each room
type) are created (the ones shown in Figure~\ref{fig:simulation_observation}).
The robot is then allowed to explore each of the two test rooms for 500,000
time steps (a duration long enough to achieve a close to uniform distribution
over the positions of the robot). A ``heat-map'' is used to visualize what
experiences the units across the representational hierarchy come to encode: it
shows the probability of the robot being in a certain position whenever a
transition sequence is observed that is encoded by a given unit, up until the
observation of the next transition sequence.

As seen in the Figure~\ref{fig:Ys_results}, lower levels of the
representational hierarchy encode experiences that are of short duration and
specific, whereas higher levels encode experiences that have longer durations
and become more abstract. In particular, on the first levels of the hierarchy,
experiences of the robot as running into a corner or moving along a wall
are represented. Also experiences that are specific to objects are encoded by
some units, as for example seen in the case of a unit on the second level,
which represents the experience of approaching two of the \emph{y}-type objects
from the top. On the third level, we already find units representing more
abstract experiences, such as ``entering a room from the corridor''. One of the
units on the fourth level seems to have already captured the experience of
roaming the \emph{s}-type room, but is not yet entirely selective: the
transition sequences that it represents are also observed in the lower part of
the \emph{y}-type test room.  On the last hierarchical level, the two units
have finally separated the experience of being in either one of the two room
types.

\section{Discussion}

The simulation experiment shows that the agent can discover sensorimotor
contexts: by exploring its environment, it learns to distinguish the two
different room types. For the agent's policy, the observed transition
probabilities of course depend on the room in which it currently is (for
example, it will never move from a wall to a \emph{y}-type object in an
\emph{s}-type room). Therefore, the agent would not have been able to learn a
good predictive model directly from the sequence of observations that it made
during its exploration. Given the learned hierarchical transition model
however, it could now separate experiences belonging to either one of the two
room types. This way, the agent has constructed an internal representation of
sensorimotor contexts, which it can use to demarginalize the observed
transition probability, and thus can reduce its prediction error when learning
a predictive model.

The work presented in this paper still leaves a number of open questions
unaddressed. For example, in the simulation experiment, the number of clusters
for each hierarchical level was manually selected. It would be desirable to
find a method to automatically determine a good number of clusters for the
construction of each new level, for example by using an Entropy-based measure.
Furthermore, from an external point of view it makes sense to distinguish two
contexts in the simulated environment: one for each room type. Consequently,
having a representational hierarchy with two states at the top level is a
suitable choice. Generally however, it seems like a mechanism is required to
determine when a good demarginalization has been already achieved, such that no
further hierarchical levels might be needed to represent the environment.

\balance

In the simulation experiment described in this paper, the agent's observation
and action spaces were discrete. It should however be noted that the presented
method can also deal with continuous inputs, by introducing a first step of
discretization. For example, observations lying in a continuous space could be
discretized using K-means (cf. \cite{laflaquiere2015developmental}).

In future work, the presented learning method for hierarchical transition
models will be combined with the training of predictive models. In such a
setting, the method presented here can be used to discover sensorimotor
contexts, such that subsequently a predictive model can be learned for each
individual sensorimotor context. This way, an agent could learn predictive
models for its own actions also in partially observable environments, where
latent environmental factors influence the outcome of the agent's actions.

\bibliographystyle{IEEEtran}
\bibliography{IEEEabrv,nhemion_ICDL-Epirob-2016}

\end{document}